\documentclass[letterpaper, 10 pt, conference]{ieeeconf}  
\IEEEoverridecommandlockouts                              
\usepackage{float}
\usepackage[normalem]{ulem}
\usepackage{amsmath}
\usepackage[noadjust]{cite}
\usepackage{siunitx}

\makeatletter
\let\NAT@parse\undefined
\makeatother
\usepackage[colorlinks,bookmarksopen,bookmarksnumbered,citecolor=red,urlcolor=red]{hyperref}
\usepackage[capitalise]{cleveref}

\usepackage{booktabs}
\usepackage{tabularx}
\usepackage{array}

\usepackage[colorinlistoftodos, disable]{todonotes}

\usepackage{comment}

\title{\LARGE \bf
GelSphere: An Omnidirectional Rolling Vision-Based Tactile Sensor for Online 3D Reconstruction and Normal Force Estimation}

\begin{document}

\hypersetup{
    colorlinks=true,
    urlcolor=blue,
    linkcolor=red,
    filecolor=magenta
}

\author{Seoyeon Lee$^{1*}$,
        Mohammad Amin Mirzaee$^{1*}$,
        and Wenzhen Yuan$^{1}$
        \\ \href{https://leahsylee.github.io/GelSphere/}{leahsylee.github.io/GelSphere}
\thanks{$*$ Equal contribution}
\thanks{
$^{1}$ are with University of Illinois at Urbana-Champaign, Champaign, IL, USA 
        {\tt\small \{leahlee2,mirzaee2,yuanwz\}@illinois.edu}}%
        }
        
\maketitle

\thispagestyle{empty}
\pagestyle{empty}


\begin{abstract}

We present GelSphere, a spherical vision-based tactile sensor designed for real-time continuous surface scanning. Unlike traditional vision-based tactile sensors that can only sense locally and are damaged when slid across surfaces, and cylindrical tactile sensors that can only roll along a fixed direction, our design enables omnidirectional rolling on surfaces. We accomplish this through our novel sensing system design, which has steel balls inside the sensor, forming a bearing layer between the gel and the rigid housing that allows rolling motion in all axes. The sensor streams tactile images through Wi-Fi, with online large-surface reconstruction capabilities. We present quantitative results for both reconstruction accuracy and image fusion performance. The results show that our sensor maintains geometric fidelity and high reconstruction accuracy even under multi-directional rolling, enabling uninterrupted surface scanning.
\end{abstract}

\vspace{5pt}

\section{Introduction} 
Surface inspection is highly demanded in industrial maintenance and manufacturing stages \cite{zheng2021recent, wang2020smart}. Defects such as scratches or cracks can compromise product safety and reliability, particularly in sectors like aerospace, automotive, and energy systems \cite{agarwal2023robotic, dastgerdi2022influence}. Beyond factory inspection, fast local shape sensing is also useful in field maintenance and in situations where a robot or a human operator must inspect curved or hard-to-reach surfaces. In medical applications, 3-D reconstruction of human skin has strong potential in dermatology and cosmetology \cite{padmanabha2025vivo}, enabling quantitative assessment of conditions such as eczema and psoriasis, monitoring treatment outcomes, and analyzing aging and sun damage.

Conventional optical inspection systems, such as structured light scanners and laser profilometers, provide high-resolution measurements, but are limited by sensitivity to ambient lighting, specular reflections, and occlusions. Furthermore, these systems often require bulky setups and cannot operate effectively in confined or curved areas \cite{salvi2004pattern, sansoni2009state, hausler2011limitations}. These challenges limit the efficiency of the aforementioned techniques in many real-world applications that require rapid, generalized surface inspection.

In contrast, GelSight-like vision-based tactile sensors (VBTS) \cite{yuan2017gelsight,johnson2009retrographic, johnson2011microgeometry, johnson2011surface} capture local surface geometry through physical contact, offering robustness to lighting and material variations. Traditional GelSight-type sensors have achieved micron-level resolution but are constrained to small areas, requiring repeated press-and-lift operations in inspection platforms \cite{agarwal2023robotic}. Rolling tactile designs such as GelBelt \cite{mirzaee2025gelbelt}, GelRoller \cite{zhang2024gelroller}, or TouchRoller \cite{cao2023touchroller} have addressed this by introducing continuous scanning, yet they remain restricted to a single axis of motion or planar surfaces due to their mechanical constraints and form factors. Besides that, wired communication that exists in most of the GelSight sensors, can limit the applicability of the sensors for rapid scanning in industrial settings.

\begin{figure}
    \centering
    \includegraphics[width=\linewidth]{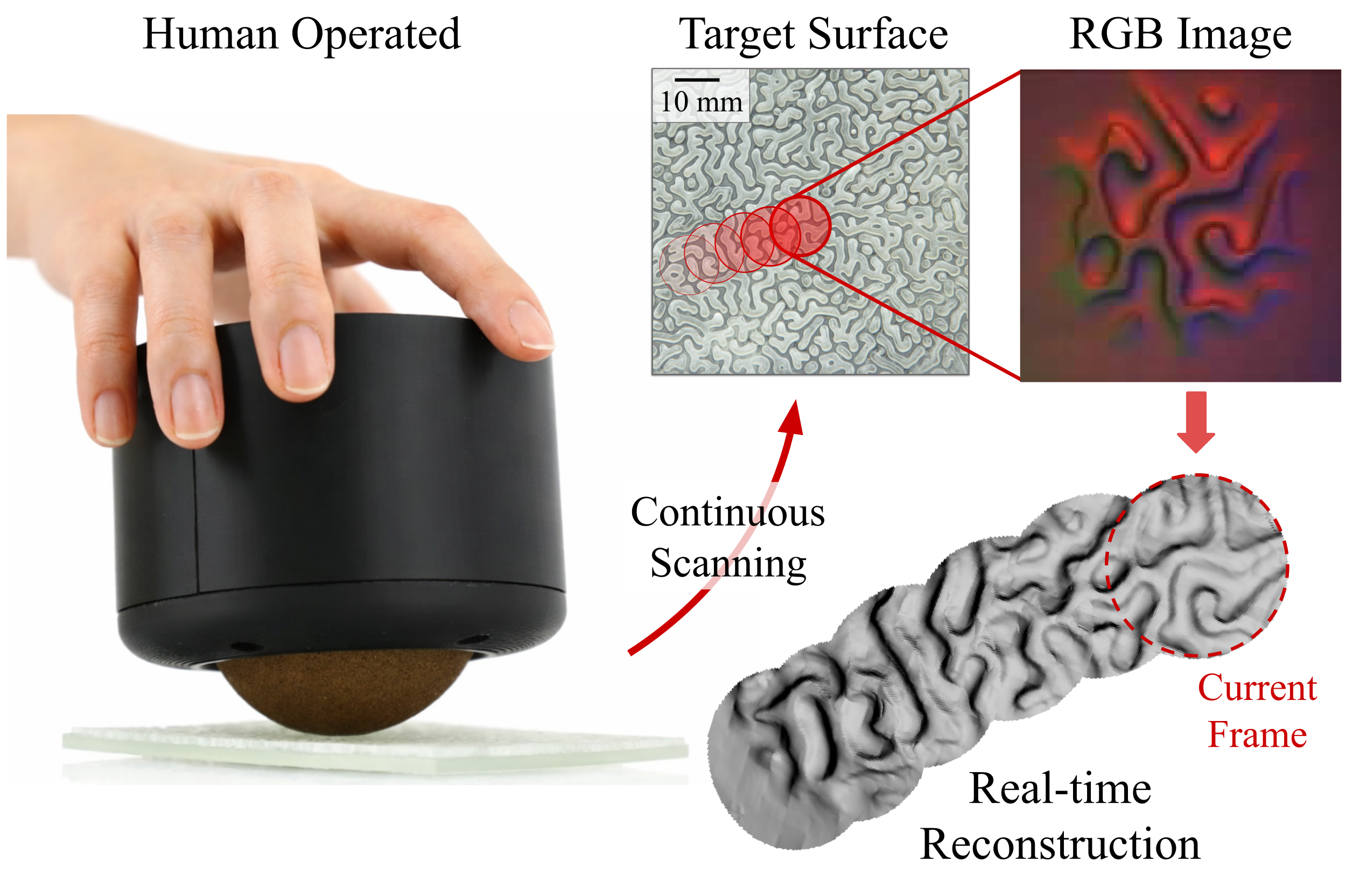}
    \caption{Our novel vision-based tactile sensor, GelSphere, achieves omnidirectional continuous scanning of large surfaces through a unique mechanical design. It can be easily human-operated for online surface reconstruction or potentially used in robotic platforms for automation. We use tactile SLAM methods to stitch the contact patches together, forming a continuous surface model.}
    \label{fig:sensor}
\end{figure}

To overcome these limitations, we propose a spherical tactile sensor that can roll freely in any direction while continuously capturing tactile images. The sensor uses a bearing-supported rolling gel surface and a magnetically stabilized internal optical module, which helps maintain a consistent camera view during motion. The system is self-contained and wireless, integrating the camera, illumination, and battery inside the housing and streaming frames over Wi-Fi. This design supports operation without cable drag and can scan large surfaces with real-time mesh reconstruction capability using GelSlam \cite{huang2025gelslam}, a real-time, high-fidelity, and robust 3D tactile SLAM system.

We present qualitative and quantitative results for both single-frame and global surface reconstruction performance of our sensor, comparing two different specular and matte surface coatings that affect the surface performance, as discussed in \cite{agarwal2025modularized}. We report spatial accuracy over the entire contact area of the sensor with a reference hex geometry. We also calculate the planar drift of the fusion method when continuously scanning a larger surface by comparing the distance and angles between specific features on the target object. Results show a planar distance drift (MAE) of $1.07$ and $1.25~\mathrm{mm}$, and angular drift (MAE) of $1.3^\circ$ and $2.7^\circ$ for specular- and matte-coated sensors while scanning a $75\times32.5\ \mathrm{mm^2}$ area.

Together, these results demonstrate that GelSphere, with its unique omnidirectional and standalone continuous scanning capabilities, can be utilized in many applications and platforms, such as surface inspection in industrial settings and skin observation in medical applications. It can be operated manually by hand or potentially be integrated into robotic platforms for automated scanning of large surfaces. The design configuration is scalable and can be adapted to different sensor sizes depending on the application. Larger versions and a higher camera frame rate can allow a higher scanning speed.

\section{Related Works}
\subsection{Vision-Based Tactile sensors}

Vision-based tactile sensors (VBTS) estimate contact geometry and contact conditions by imaging the deformation of a soft optical interface.
GelSight \cite{yuan2017gelsight,johnson2009retrographic, johnson2011microgeometry, johnson2011surface}, made VBTS practical by introducing a compact optical stack with multi-directional lighting and a reflective membrane on top of a soft elastomer. 
An internal camera captures surface deformations, enabling dense 3D reconstruction, shear estimation, and slip detection \cite{yuan2015measurement, dong2017improved, zhang2026fruittouch}. GelSight-like sensors have since been widely adopted for robotic manipulation tasks, including grasp stabilization, object localization, and in-hand regrasping \cite{li2014localization, calandra2018more}.

A large body of work has since focused on changing the sensor form factor and improving robustness for integration into robotic end-effectors. Examples include GelSlim \cite{gelslim1} and GelSlim 3.0 \cite{taylor2021gelslim3}, FruitTouch \cite{zhang2026fruittouch}, and other optical tactile sensor families such as TacTip and DenseTact \cite{ward2018tactip, lepora2021soft, 9811966, do2023densetact, do2023inter, mirzaee2024multiphysics}. Other studies explored different shapes and larger sensing coverage, including GelSight Svelte \cite{zhao2023gelsight}, GelSight Fin Ray variants \cite{finrayliu2022gelsight, babyfinray}, and hand/finger systems with embedded vision-based tactile sensing \cite{sandraExoGripper, sandraEndoFlex}. More recent curved and omnidirectional tactile sensors, such as GelSight360 \cite{tippur2023gelsight360} and RainbowSight \cite{tippur2024rainbowsight}, show that non-planar optical tactile surfaces can increase sensing coverage and improve shape reconstruction on complex contacts.

Most of the systems above are designed for local sensing in robotic manipulation setups and are limited in surface coverage. They require repeated contacts and sequential probing to explore larger areas \cite{zhang2025pneugelsight, agarwal2023robotic}, which introduces discontinuities in measurement and increases reconstruction time. Such lift-and-recontact strategies are well-suited for grasp refinement or localized inspection, but they are inefficient for continuous surface exploration or large-area geometry reconstruction. 

Our work instead targets continuous scanning over larger surfaces while retaining high-resolution GelSight sensing.

\subsection{Continuous Surface Scanning}

Several recent works address large-area sensing by improving the mechanical design so that the elastomeric sensing material can roll across the surface. Cao et al. use a rolling cylindrical design for 2D surface texture measurement over large areas \cite{cao2023touchroller}. Other cylindrical visual-tactile sensors improve information quality during motion using image fusion or enhanced processing for 3D reconstruction of the surface \cite{li2023enhanced, zhang2024rotip}. Roller-based manipulation systems also show the practicality of rolling contact for increasing contact coverage without repeatedly lifting and regrasping \cite{yuan2023tactile, yuan2022robot}. Cylindrical sensors overcome the limitation of the traditional sensors in continuous scanning, yet their performance and speed are limited due to a narrow sensing area. GelBelt uses a belt of coated elastomer as the sensing material for continuous sensing and reconstructs large areas at a higher speed, 45 $mm/s$ and precision \cite{mirzaee2025gelbelt}. 

The above sensors motivate the use of rolling tactile interfaces for efficient scanning, but all of them are constrained to 1D motion on the surface, which limits the sensing capability and generalizability to different surfaces and geometries. 

In addition to the hardware design, a key challenge in large-surface tactile inspection is that each tactile image only covers a small local contact patch. To reconstruct a larger region, the system must estimate the motion of the sensor between frames and fuse many local measurements into one global map. 
Optical-flow-based methods are commonly used when the tactile image contains enough local texture and inter-frame motion is small \cite{Lucas1981}. ICP is also studied for tracking the motion of the contact object using the reconstructed mesh of the indented area \cite{gelwedge}. In other settings, external pose information from robot kinematics or tracking systems can be used to support stitching, as in robot-mounted tactile scanning systems \cite{zhang2025pneugelsight, agarwal2023robotic}. 

Recent Tactile mapping and tactile SLAM methods like GelSlam and NormalFlow \cite{huang2025gelslam, huang2024normalflow} align repeated local features in the curvature map through optimization algorithms to recover large object geometry and surface structure. We employ this technique for real-time reconstruction in our sensor.
\begin{figure}
    \centering
    \includegraphics[width=\linewidth]{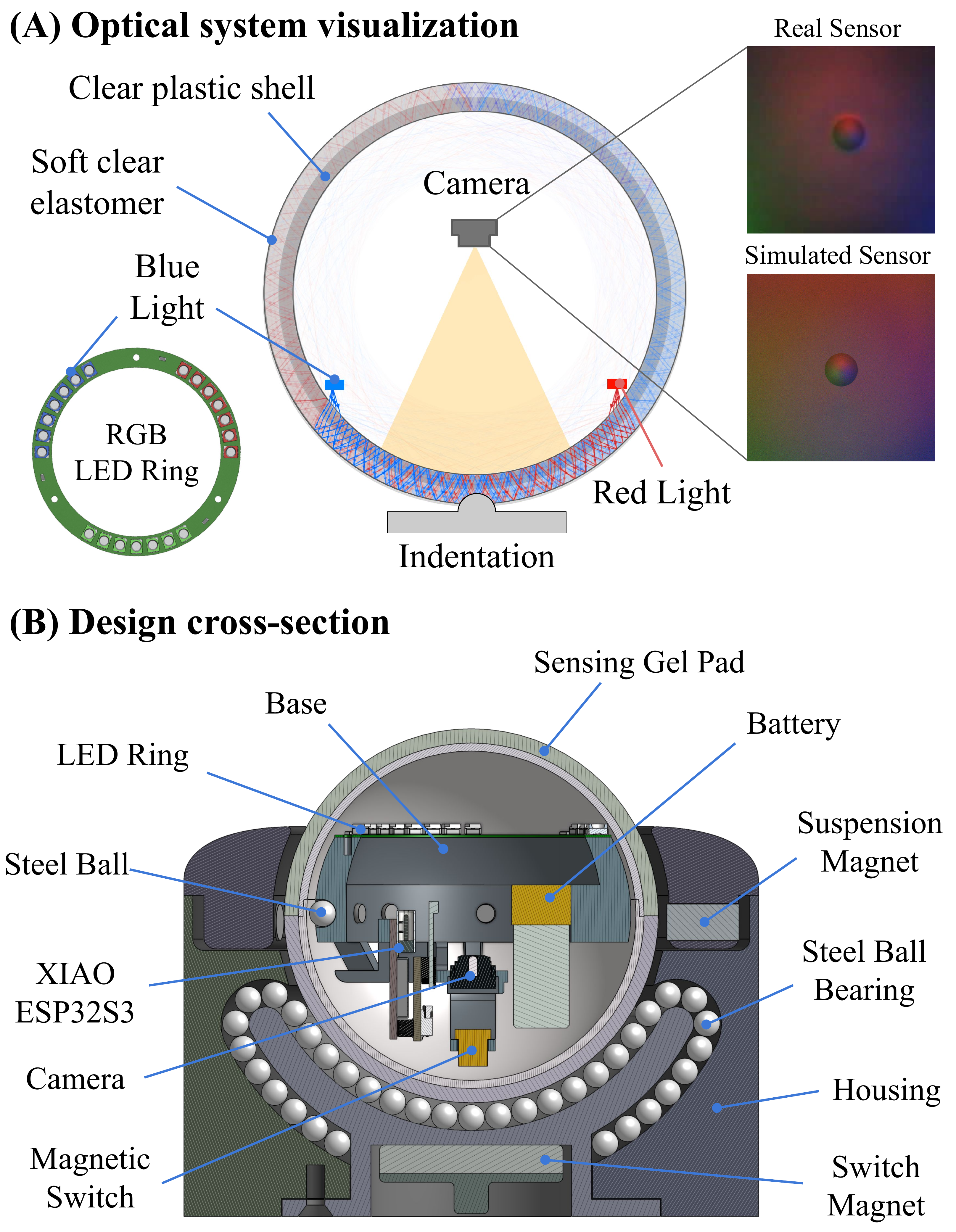}
    \caption{Optical and mechanical design of GelSphere. (A) The simulated tactile image is shown alongside a real sensor image under a spherical indentation, presenting a close visual similarities between the optical simulation and the fabricated sensor. The RGB LED ring consists of 18 LEDs (6 of each color), and three color groups are spaced by approximately $120^\circ$, which provides directional illumination from every direction. (B)  Cross-sectional view of the GelSphere sensor. The internal optical module (camera, LED right, and battery) is mounted on a base that is magnetically suspended, which helps maintain a consistent camera orientation while the outer shell rolls. A magnetic switch enables power on/off control without opening the housing. The steel ball bearing layer ensures low-friction contact between the sensing gel assembly and the housing, enabling smooth rolling during scanning. The sensor is self-contained, with onboard power and wireless image streaming. }
    \label{fig:sensordesign}
\end{figure}

\section{Sensor Design and Fabrication}

\begin{figure*}
    \centering
    \includegraphics[width=\linewidth]{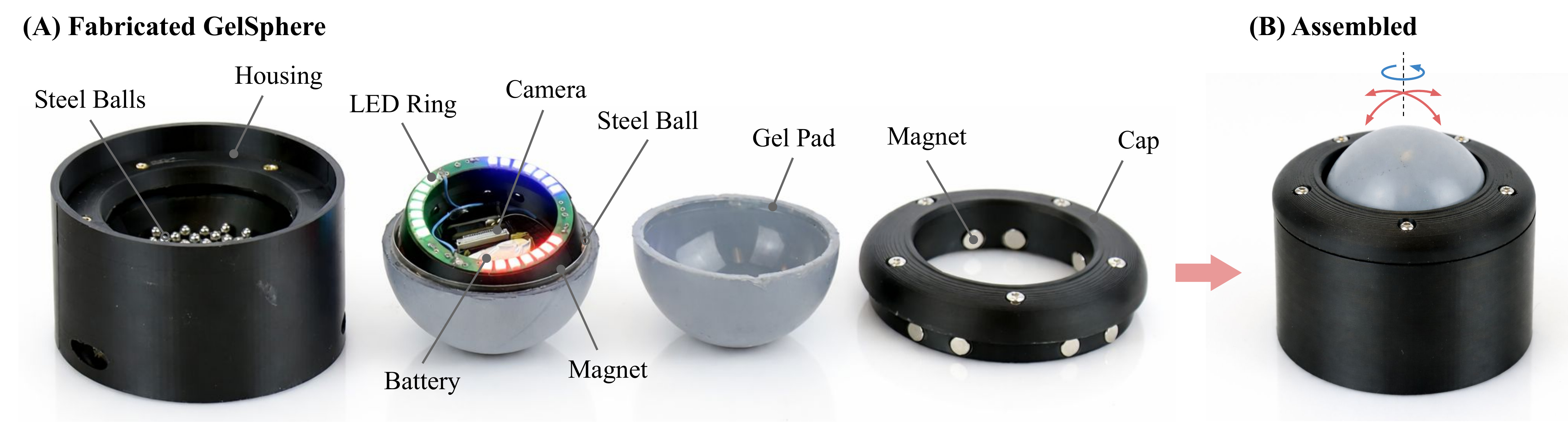}
    \caption{GelSphere sensor design. (A) Fabricated GelSphere components. The sensing gel pad is removable to allow battery replacement and maintenance. The cap is fastened to the housing to support the magnetic suspension structure and retain the steel balls in the bearing race. (B) Assembled prototype. The sensor is able to roll omnidirectionally when fully assembled. }
    \label{fig:mechanical}
\end{figure*}

\subsection{Sensor Design}

Our sensor features a spherical sensing gel supported by an inner acrylic shell and enclosed in a rigid PLA-printed housing, as illustrated in \cref{fig:sensordesign}. The mechanical design is inspired by a ball-transfer unit, where miniature steel balls form a bearing layer between the gel sphere and the housing surface, substantially reducing friction and allowing the sphere to roll smoothly in any direction. 

Inside the sphere, a compact camera module with Wi-Fi communication, a circular RGB LED ring, and a battery module are mounted on a lightweight optical base. The spherical shell is designed in two hemispheres to allow direct access to the internal electronics for maintenance, such as battery recharging or replacement. This design also allows convenient replacement of a damaged gel pad, because only the affected hemisphere needs to be replaced. 

The optical base is suspended magnetically through coupling magnets around the housing and the base on the midplane level of the sphere.This coupling passively constrains the orientation of the base, ensuring the camera faces the sensing surface regardless of the orientation of the sensor. 

The optical design follows the standard principles of the GelSight family of sensors, which includes a clear soft elastomer over a clear rigid shell, while the outer surface of the elastomer is coated with a reflective material. In our sensor, an RGB LED ring is mounted around the camera to illuminate the gel at a shallow angle. We conducted a physics-based simulation \cite{agarwal2025modularized} to adjust the location of the LEDs in the sensor. The optical system visualization and the simulated image of the sensor are illustrated in \cref{fig:sensordesign}A. We observe that the simulated result highly matches the real sensor output.

Similar to other GelSigh-like sensors, the clear gel’s exterior is coated with a reflective layer that encodes surface deformation into light-intensity patterns visible to the camera.
Generally, a diffused Lambertian coating is ideal for perfect photometric stereo systems; however, studies have shown that specular coatings can improve the performance in curved GelSight surfaces, such as a fingertip sensor \cite{agarwal2025vision}. To investigate the specularity effect in our sensor, we fabricate two different sets of gel pads with coatings of different specularities and compare the reconstruction results.

An embedded magnetic switch is integrated into the power circuit, allowing activation or shutdown without physical openings of the sphere. \Cref{fig:sensordesign} illustrates the cross-section view of the GelSphere and its internal components. The overall sensor dimension measures approximately $100~\mathrm{mm} \times 100~\mathrm{mm} \times 90~\mathrm{mm}$ with the sensing gel diameter of $65.3~\mathrm{mm}$.

The spherical form factor provides a uniform contact footprint regardless of motion direction, making it well-suited for handheld or robotic inspection of large surfaces.

\subsection{Fabrication}

\begin{figure}
    \centering
    \vspace{10pt}
    \includegraphics[width=\linewidth]{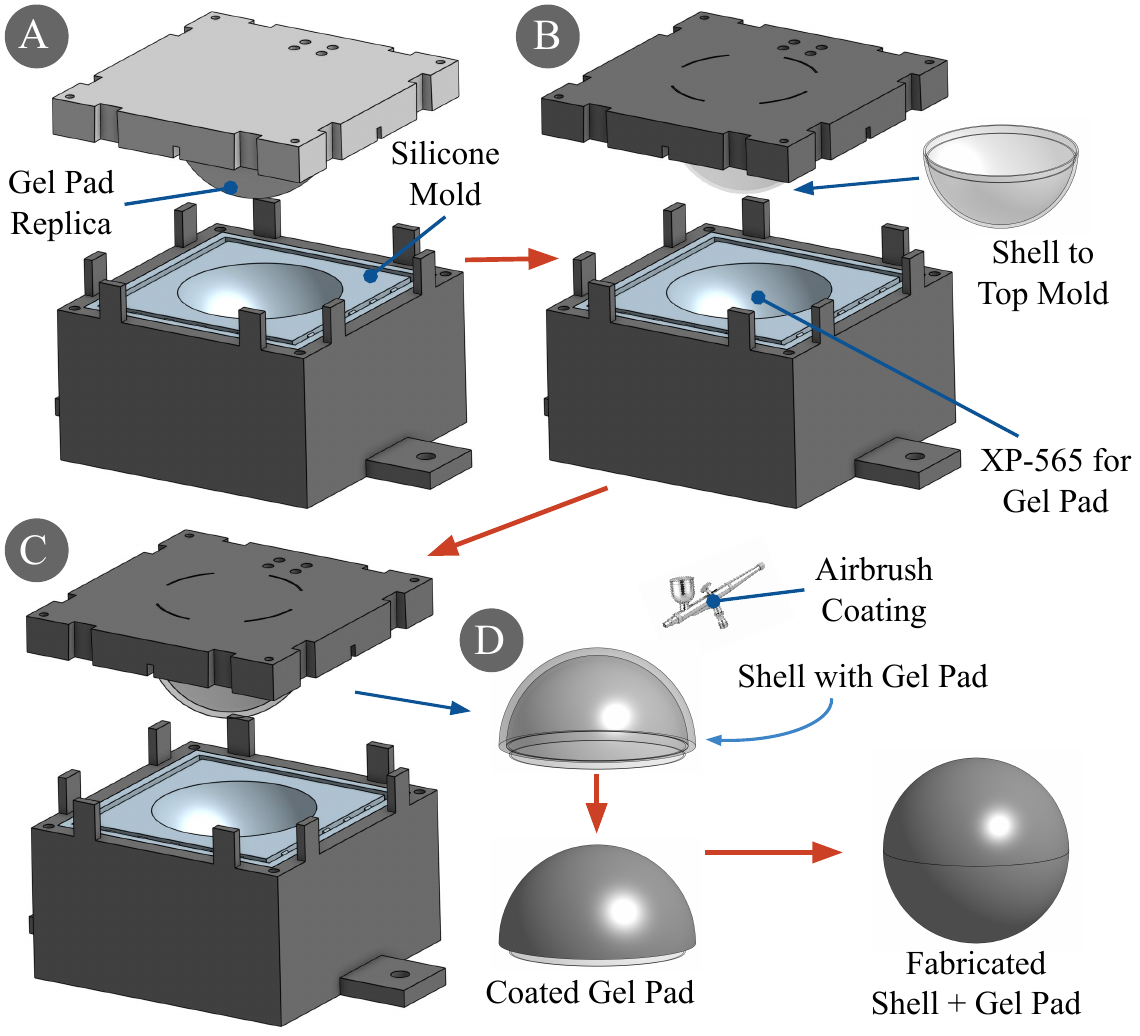}
    \caption{Fabrication Process for making the gel pad and the shell. The silicone mold (Mold Star 15) is created using the gel pad replica, and XP-565 silicone elastomer is added to the cured silicone mold. The off-the-shelf plastic shell is added to the top mold and pressed carefully with the bottom part of the mold, ensuring no air is caught. The cured elastomer is air-sprayed with the desired coating material to finish the gel pad.}
    \label{fig: fabrication}
\end{figure}

The tactile gel is fabricated as two hemispherical shells, as shown in \cref{fig: fabrication}. We first create a silicone mold of the target gel pad geometry using Smooth-On Mold Star 15. To make this mold, a replica of the desired gel shape is SLA-printed (Clear resin V4, Formlabs) and polished to obtain a smooth surface finish. This gel pad replica is sprayed with Mann Ease Release 200 and gently pressed onto the Mold Star silicone that was poured into a PLA-printed container. After the mold cures, Mann Ease Release 200 is sprayed again onto the mold cavity to prevent adhesion. An off-the-shelf clear hemispherical shell is placed on a PLA-printed holder and used as the gelpad supporting membrane. We then press it into the silicone elastomer (Silicone Inc.\ XP-565, mixing ratio $1{:}16$ for parts A:B) that was poured into the mold. After curing, the two halves can be assembled together to form a complete sphere. 

We fabricate two coating variants for reconstruction performance comparison. For the matte sensing surface, we first prepare a silicone ink by mixing black ink, white ink, and thinner in a $1{:}6{:}21$ ratio (Print-On Silicone Ink, Raw Material Suppliers), and then combine this mixture with silicone part B in a $40{:}1$ ratio. The resulting ink is airbrushed onto the cured gel surface to form a diffusive reflective layer. For the specular coating, we mix bronze powder with Novocs (Smooth-On) and XP-565 silicone and airbrush the mixture onto the surface. The specular coating produces higher-contrast tactile images.

The inner base and the outer housing of the sensor are 3D-printed in PLA, and 4.5 mm diameter steel balls are used to create a bearing layer for smooth rolling motion. 

The camera we use is Seeed Studio XIAO ESP32-S3, which supports wireless connection through Wi-Fi, allowing real-time streaming to other devices. The camera systems, including the boards, have small dimensions of $17.8~\mathrm{mm} \times 21~\mathrm{mm} \times 15 ~\mathrm{mm}$, making it suitable for integration inside the spherical shell. Although the module supports streming at higher resolutions, we stream the frame with a size of $240 \times 240$ pixels to balance spatial resolution and latency. We set the exposure time at 100 ms and fixed other settings to capture consistent frames.

The magnetic switch is placed at the top of the optical base, interacting with a neodymium disc magnet outside the housing. By default, the switch is on, and placing the magnet cuts the power going through the board from a small 3.7 V lithium-Ion battery.
\Cref{fig:mechanical} shows the fabricated sensor and its components. 

\section{Geometry and Force Measurement}

In this section, we describe how the tactile images captured by GelSphere are converted into local surface geometry, large-area surface reconstructions, and contact force estimations. As the sensor rolls over textured 3D surfaces, it captures tactile images, which we use to estimate the local surface geometry. Then, we determine the sensor's frame-to-frame planar movement and compose the local geometries into a global 3D shape. We also use the tactile data to estimate the normal contact force applied to the sensor.

\subsection{Geometry Reconstruction from Single Frame}
The GelSphere uses the photometric stereo method, like other GelSight-like sensors, to estimate the surface normal. Our calibration method is similar to \cite{gelwedge}. We press spherical indenters with diameters of 6 mm, 7 mm, and 8 mm, each on 20 different areas of the sensor. We then manually label the contact regions in the tactile images. The resulting dataset is used to train a lightweight neural network that maps the pixel color intensity change and position to local surface gradients. For the calibration stage, we assume a flat background surface; however, in the test stage, we transform the estimated normal map to the sensor's spherical surface curvature for normal map correction.

We use common objects to qualitatively show the reconstruction performance and press a known hex shape on different locations of the sensing area to quantify the spatial reconstruction accuracy of the sensor, similar to \cite{mirzaee2025gelbelt}. 

\subsection{Global Surface Reconstruction}
During continuous rolling, our sensor produces a sequence of tactile images with overlapping contact regions. To estimate the contact region, we assume the sensor’s spherical surface forms an approximately circular contact patch within a radius range covering the majority of the tactile signals in the subtracted background image. Each frame is first converted into a dense normal map using the calibrated model. Our goal is to fuse these local measurements into a single, globally consistent mesh.

We use the method and codebase provided by GelSLAM \cite{huang2025gelslam} to fuse the contact patches together. This method uses the normal and curvature maps similar to \cite{huang2024normalflow} for frame-to-frame pose estimation and local transformation in the tactile stream before performing the loop closure steps consisting of SIFT matching and pose graph optimization on the contact patches of all frames. Finally, the global surface is reconstructed by registering the point cloud of the contact patches to form a fused surface.

\subsection{Contact Force Estimation}
While vision-based tactile sensors directly measure the deformation of the gel upon contact, the contact force can be inferred based on the deformation. Compared to flat sensors, the spherical geometry of our sensor enhances the RGB variation to different contact forces. This is applicable even on a flat object, because increasing the force results in a larger deformed area in the image.

We use a machine-learning method to directly estimate the contact normal force from raw sensor images. We collect data on normal forces ranging from 1 to 15 N using eight indenters with varying surface curvatures, from 1 to 64 cm in powers of two, and on a flat surface. We train a 4-layer CNN model on the original $240\times240$ RGB images of the sensor to directly estimate the normal force.

\begin{table*}[t]
\vspace{15pt}
\centering
\caption{Comparison of reconstruction performance under specular (bronze) and matte coatings for single-frame and large-surface reconstruction. Higher Dot Product, SSIM, and PSNR, and lower MSE/MAE indicate better alignment with ground truth.}
\label{tab:coating_comparison}

\begin{tabularx}{\linewidth}{>{\bfseries}l c c c c c c c c c}
\toprule
& \multicolumn{5}{c}{(A) Single-frame (Hex Indenter)} & \multicolumn{4}{c}{(B) Large-surface (Feather)} \\
\cmidrule(lr){2-6} \cmidrule(lr){7-10}
Coating & Dot Product & MSE & MAE & SSIM & PSNR [dB] & Dist. MSE [$mm^2$] & Dist. MAE [$mm$] & Angle MSE [$^{\circ2}$] & Angle MAE [$^\circ$] \\
\midrule
Specular & 0.965 & 0.023 & 0.111 & 0.950 & 22.62 & 1.6169 & 1.070 & 3.7 & 1.3 \\
Matte    & 0.939 & 0.041 & 0.156 & 0.930 & 20.08 & 2.1813 & 1.251 & 12.8 & 2.7 \\
\bottomrule
\end{tabularx}
\end{table*}
\begin{figure}
    \centering
    \includegraphics[width=\linewidth]{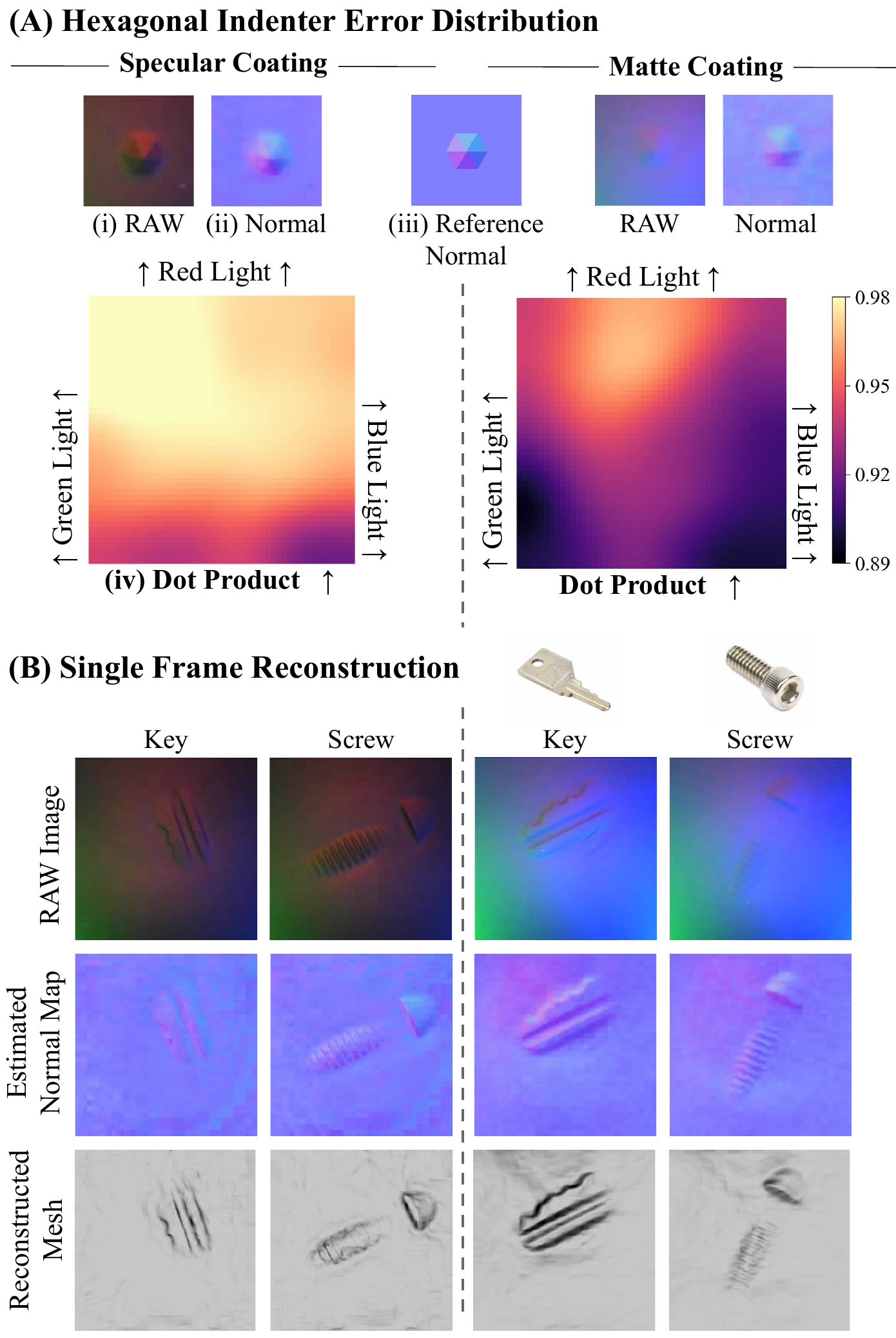}
    \caption{GelSphere's single frame sensing performance. (A) A 2D normal-estimation accuracy plot was generated from the indentation of a hexagonal pyramid at multiple locations. The RGB direction labels in the dot-product maps are for visual reference. The three LED groups are positioned $120^\circ$ apart around the ring. Left: specular, Right: matte. (B) A key and an M8 screw were pressed into both specular and matte sensing gel, showcasing the reconstruction performance.}
    \label{fig: single_and_hex}
\end{figure}

\section{Experiments}

The GelSphere enables continuous surface scanning in all directions, measures miniature details, and stitches contact patches together. We evaluate these capabilities in three sets of experiments: single-frame reconstruction with small objects and evaluating the 3D reconstruction performance, reconstruction of larger surfaces by the continuous movement of the sensor, and contact normal force estimation. 

\subsection{Single-frame Reconstruction}

We first demonstrate the qualitative sensing capability by pressing small objects, such as a screw and a key, on the sensing surface. \Cref{fig: single_and_hex}B shows the raw tactile image, the estimated surface normal map, and the reconstructed 3D mesh of the objects pressed on both specular and matte gel pad. Both gel pads demonstrated detailed surface geometry, such as the thread on the screw and dimples in the key.  

To quantitatively evaluate the accuracy of our surface normal estimation, we press a hexagonal pyramid indenter with known geometry at 50 different locations over the sensing area of the sphere. \Cref{fig: single_and_hex} Ai shows the raw tactile image of a hex indenter pressed on the sensing gel, and Aii shows the estimated surface normal from the same image, while Aiii presents an example reference normal from a hex indent. For each test image, we manually mask the hex indentation and compare it with the reference normal map. We then calculate the average dot product of the reconstructed and ground truth normals in the masked hex shape, as illustrated in \cref{fig: single_and_hex}Aiv. We observe that the areas closer to the red light have a higher accuracy, while the pixels closer to the blue light result in a higher normal estimation error. This error distribution can potentially be the result of the high-intensity blue light that gets saturated as we get closer to the blue LEDs. Although the accuracy range of the two coatings is almost the same, we observe that the specular coating provides much more consistent accuracy over the surface, which matches the results shown in \cite{agarwal2025vision}, suggesting that in a specular setting, illumination quality is better preserved through the curved gel pad geometries.

Table \ref{tab:coating_comparison}A shows the computed values for the mean square error (MSE) and mean absolute error (MAE) between the estimated and reference normals. We also treat the normal fields as RGB images and compute the structural similarity index (SSIM) and peak signal-to-noise ratio (PSNR).

\subsection{Global Surface Reconstruction}

\begin{figure}
    \centering
    \includegraphics[width=\linewidth]{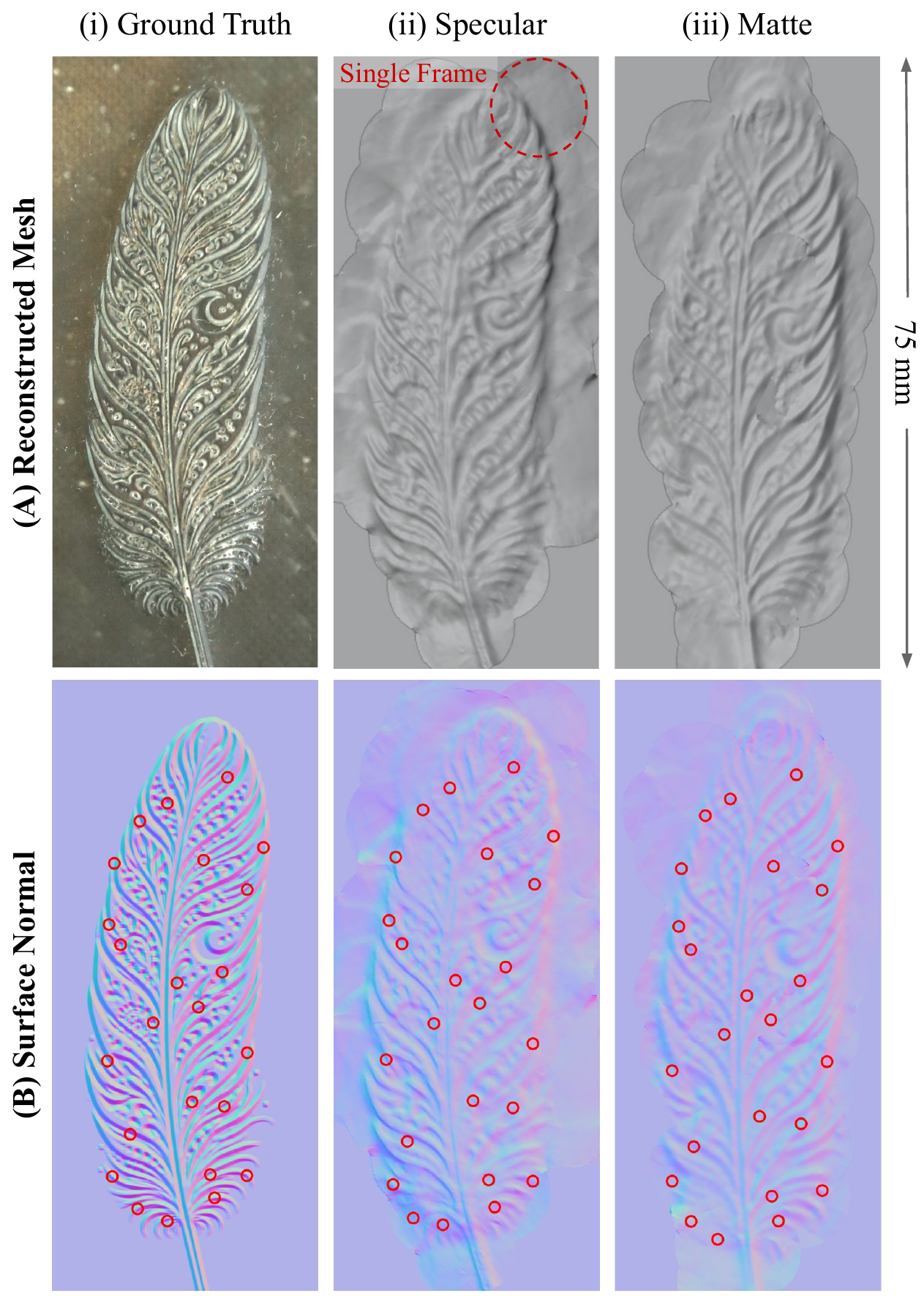}
    \caption{Large surface reconstruction of an SLA-printed feather using specular and matte-coated GelSphere sensors. (A) The printed ground-truth and the reconstructed meshes from specular and matte coatings. (B) Surface normal maps are used for error analysis. 24 correspondence points are manually selected on the ground truth and each reconstructed normal map. These points are used to compare pair-wise distances and local normal directions between corresponding points. The specular coating yields lower error than the matte coating, as shown in \cref{tab:coating_comparison}.}

    \label{fig: large_surface}
\end{figure}

We test the large-surface reconstruction pipeline and accuracy by rolling the sensor over an object of known geometry. We SLA print (Clear resin V4, Formlabs) a high-resolution feather pattern of dimension $75~\mathrm{mm} \times 32.5~\mathrm{mm}$ on a flat surface, as shown in \cref{fig: large_surface}A (i).  We then roll GelSphere over the entire textured surface manually while observing the scanned areas in GelSLAM's
real-time view window. We further flattened the reconstructed mesh shown in \cref{fig: large_surface}A. We observe that both specular and matte coatings provide detailed information about the surface, yet the specular version has preserved more detail and geometry consistency.

In \cref{fig: large_surface}B, we perform a quantitative study on the planar drift of the stitching method. We randomly choose 24 reference points (marked in red) on the normal map of the original mesh and find the correspondence in the reconstructed normal maps. We then use the line segments formed by the combinations of all the points to calculate the distance and angular errors.  \Cref{tab:coating_comparison}B shows the calculated MSE and MAE in the distance and the local normal directions between corresponding points. The specular coating results in lower error both in distance and angle. 

\subsection{Contact Force Estimation}

\begin{figure}
    \centering
    \includegraphics[width=\linewidth]{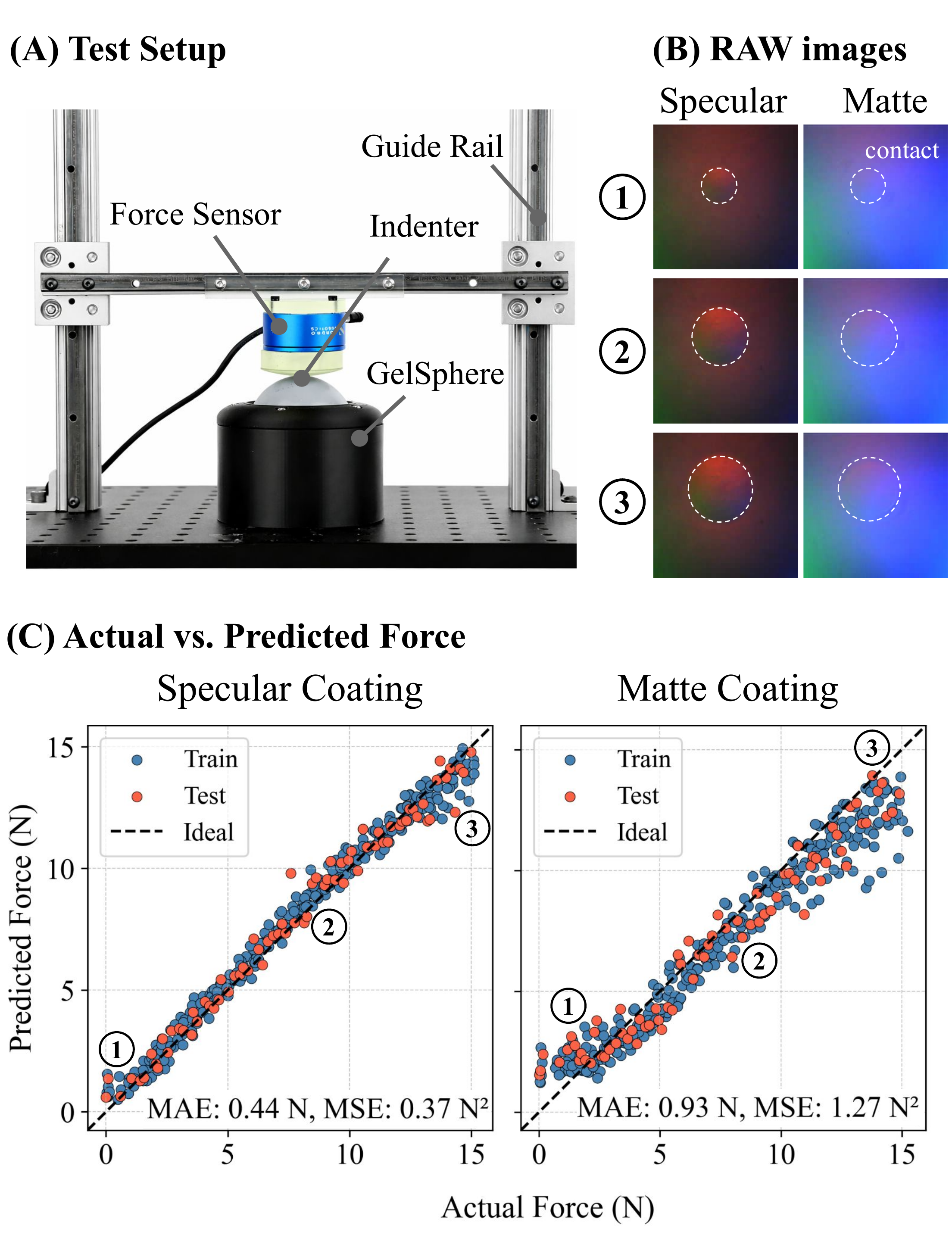}
    \vspace{-15pt}
    \caption{Contact normal force estimation using GelSphere. (A) We use linear guides to constrain motion to the z-axis in the force application setup. The ground truth force is measured using a force-torque sensor (NRS-6050-D50, Nordbo Robotics), while indenters of different radii are manually pushed onto the sensor surface. (B) The size of the contact area increases as the force increases. The specular surface has a much higher visual sensitivity compared to the matte coating. (C) The specular-coated sensor estimate forces with MAE of 0.44 N while the Matte coating has almost double the amount, resulting from a lower signal-to-noise ratio compared to specular.}
    \label{fig:force}
\end{figure}

We collected around 750 data points using the setup shown in \cref{fig:force}A, including a force-torque sensor and indenters of different radii. Eight indenters were 3D-printed using an SLA printer (Clear Resin V4, Formlabs). As expected, increasing the force enlarges the contact area as shown in \cref{fig:force}B, while the curvature information is embedded in the pixel-level RGB variation within the contact patch. Two CNN models for two different materials were trained on the raw images with an 80:20 percent train-test split, outputting the normal force.

As shown in the \cref{fig:force}C, the models were able to estimate the normal forces with a mean absolute error of 0.44 and 0.93 N and a mean squared error of 0.37 and 1.27 N on the test data for specular and matte materials, respectively. We observe that the specular material has a lower estimation error compared to the matte coating, especially in the lower forces. This can be the result of the higher surface reflectivity and sensitivity to small deformations, increasing the signal-to-noise ratio in the tactile image, which consequently improves the quality of the data fed to the model. The different sensitivities are apparent in the \cref{fig:force}B, where the pixel intensity variations in the specular-coated sensor are even visually higher than in the matte.  

\section{Conclusion}

We presented GelSphere, a spherical GelSight-like VBTS for continuous surface scanning. Compared to other GelSight derivatives that mainly provide local sensing or are mechanically constrained to a single rolling direction, GelSphere enables omnidirectional rolling contact and can potentially follow curved surfaces and irregular scan paths. The self-contained design integrates the camera, LED, and battery inside the sensor and supports wireless image streaming for both real-time and offline reconstruction.

In the single-frame hex-indenter experiment, the sensor showed high normal-estimation accuracy, with mean dot product values up to $0.98$ for the specular and $0.97$ for the matte coating. The dot product map showed better results near the red LED region and lower accuracy near the blue LED region. This suggests that further adjusting the illumination intensity could improve sensing quality. In the large-surface reconstruction experiment, the specular coating reported lower planar drift than the matte coating, with distance MAE values of $1.07~\mathrm{mm}$ and $1.25~\mathrm{mm}$, respectively. In the force-estimation experiment across surfaces with different curvatures, the specular coating also produced lower error, with MAE of $0.44~\mathrm{N}$ compared with $0.93~\mathrm{N}$
for the matte coating.

The current sensor also has several limitations. Fast scanning can reduce reconstruction quality and may cause the steel balls to get caught between the housing and the gel pad. In addition, increasing camera resolution significantly increases streaming latency.

The sensor design is also scalable and can be adapted to different sizes depending on the target application. A larger sensor could be used for faster coverage during scanning of large surfaces, while a smaller sensor could improve access to small spaces and high-curvature regions. Miniaturization is possible by using smaller batteries and electronics. However, in our current evaluation, wireless camera modules smaller than the XIAO ESP32S3 Sense showed substantial trade-offs in image resolution and/or streaming latency. Ultimately, changing the sensor size does not fundamentally alter the rolling mechanism or the core optical design.

GelSphere can be used in both handheld and robot-mounted operations. The real-time reconstruction view provides immediate feedback to a human operator during scanning, while the self-contained wireless design also makes the sensor suitable for automated inspection with motorized systems or robot-arm deployment. 


\addtolength{\textheight}{-0cm}   
\bibliographystyle{IEEEtran}
\bibliography{IEEEabrv, ref}

\end{document}